\documentclass[10pt,twocolumn,letterpaper]{article}

\usepackage[pagenumbers]{cvpr} 

\definecolor{cvprblue}{rgb}{0.21,0.49,0.74}
\usepackage[breaklinks,colorlinks,allcolors=cvprblue]{hyperref}
\usepackage{xcolor}         
\usepackage{multirow}
\usepackage{graphicx}
\usepackage{amssymb}
\usepackage{makecell}
\usepackage[table]{xcolor}
\usepackage[most]{tcolorbox}
\usepackage{xcolor}
\usepackage{marvosym}
\usepackage{fontawesome5}

\makeatletter
\newcommand\blfootnote[1]{%
  \begingroup
  \renewcommand\thefootnote{}\footnote{#1}%
  \addtocounter{footnote}{-1}%
  \endgroup
}
\makeatother

\def\paperID{7058} 
\def\confName{CVPR}
\def\confYear{2025}

\title{LatBot: Distilling Universal Latent Actions for Vision-Language-Action Models}

\author{Zuolei Li$^{1,2\dagger}$ \quad \text{Xingyu Gao}$^{1,2}$\textsuperscript{\Letter} \quad \text{Xiaofan Wang}$^{1,2\dagger}$ \quad \text{Jianlong Fu}$^{3}$ \\[3pt]
$^1$~\text{Institute of Microelectronics, Chinese Academy of Sciences} \\ $^2$~\text{University of Chinese Academy of Sciences} \quad
$^3$~\text{Microsoft Research}  \\
{\tt\small \{lizuolei24, wangxiaofan24\}@ime.ac.cn, gxy9910@gmail.com, jianf@microsoft.com} \\
\noindent\textbf{Project Page: }
\href{https://mm-robot.github.io/distill_latent_action}{
    \textcolor{blue}{\faRobot\; mm-robot/LatBot}
}
}

\begin{document}
\maketitle

\blfootnote{\textsuperscript{\text{\Letter}} Corresponding author: gxy9910@gmail.com}
\blfootnote{$\dagger$ Work conducted during internship at Microsoft Research}

\begin{abstract}
Learning transferable latent actions from large-scale object manipulation videos can significantly enhance generalization in downstream robotics tasks, as such representations are agnostic to different robot embodiments. Existing approaches primarily rely on visual reconstruction objectives while neglecting physical priors, leading to sub-optimal performance in learning universal representations. To address these challenges, we propose a Universal Latent Action Learning framework that takes task instructions and multiple frames as inputs, and optimizes both future frame reconstruction and action sequence prediction. Unlike prior works, incorporating action predictions (e.g., gripper or hand trajectories and orientations) allows the model to capture richer physical priors such as real-world distances and orientations, thereby enabling seamless transferability to downstream tasks. We further decompose the latent actions into learnable motion and scene tokens to distinguish the robot’s active movements from environmental changes, thus filtering out irrelevant dynamics. By distilling the learned latent actions into the latest VLA models, we achieve strong performance across both simulated (SIMPLER and LIBERO) and real-world robot settings. 
Notably, with only 10 real-world trajectories per task collected on a Franka robot, our approach successfully completes all five challenging tasks, demonstrating strong few-shot transferability in robotic manipulation.

\end{abstract}    
\section{Introduction}
\label{sec:intro}


Latent action learning has recently emerged as a promising research direction in the field of vision-language-action (VLA) models \cite{kim2024openvla, black2024pi_0, zhou2025vision, qu2025spatialvla, liu2024rdt, team2024octo, zitkovich2023rt}. Its core idea is to extract and compress motion semantics between consecutive frames into compact latent representations that are agnostic to different robot embodiments. Unlike traditional methods \cite{kim2024openvla, black2024pi_0, pertsch2025fast} that rely on annotated action data, this paradigm enables models to leverage large-scale human videos, thereby significantly expanding the available training sources for robotic policies and overcoming the limitations of conventional robot datasets in terms of diversity and generalization.
\begin{figure*}
    \centering
    \includegraphics[width=1\linewidth]{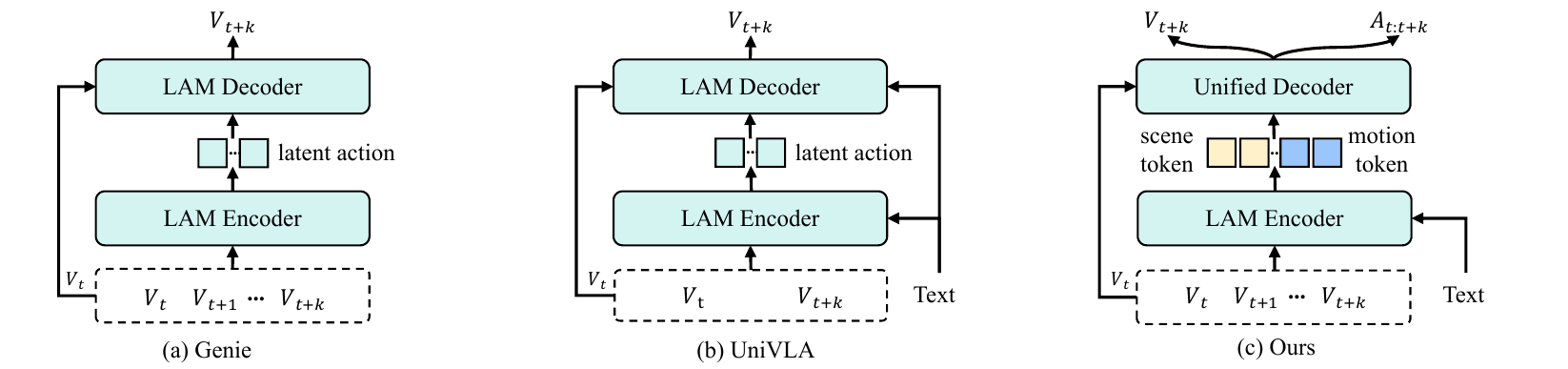}
    \caption{Different paradigms in latent action modeling (LAM). Existing methods often ignore disentangling robot actions from environmental changes. In contrast, we learn a disentangled representation and decode latent actions into both the future visual frame $V_{t+k}$ and physical actions $A_{t:t+k}$ that enables more accurate and transferable control for downstream tasks.}
    \label{fig:comparison}
\end{figure*}


However, as shown in Figure \ref{fig:comparison}, existing latent action models (LAM) \cite{bu2025univla, chen2024moto, ye2024latent, bu2025agibot}  usually suffer from the following challenges. First, the absence of task instruction guidance prevents the latent action from capturing task-relevant changes (e.g., Genie \cite{bruce2024genie}). Second, insufficient utilization of multiple frames results in imprecise latent action representations incapable of accurately capturing motion dynamics (e.g., UniVLA \cite{bu2025univla}). Third, the latent actions often focus on visual appearance changes but lacking physical awareness, causing a semantic gap between latent action representations and real executable actions. These limitations hinder the effective transfer of the learned latent actions to downstream tasks, as they fail to provide reliable cues for planning, limiting their ability to generalize from visual perception to real-world robotic execution.



To address these issues, we propose \textbf{LatBot}, a universal latent action learning framework for robotic tasks, which learns latent actions under the guidance of task instructions and multi-frame inputs, constrained by both visual and action generation objectives. This design enables VLA models to generalize more effectively across downstream tasks with few-shot samples (e.g., 25 demonstrations used in existing works \cite{ren2025motion}). First, we design two complementary learnable latent action tokens: \textit{scene tokens} to capture passive environmental changes such as object position, pose, and background dynamics, and \textit{motion tokens} to encode the robot’s active movements such as end-effector translation, rotation, and gripper actions. This design explicitly disentangles environmental variations from robot-induced motion, leading to more structured latent actions that improve motion understanding and action prediction. Second, we propose a unified decoder that conditions on the latent actions to jointly guide future frame reconstruction and inter-frame action generation. It enables the model to learn universal latent actions that not only enhance the prediction of diverse real-world robotic manipulation scenes, but also bridge the gap with real actions, ultimately improving transferability to downstream manipulation tasks. To optimize latent action representations, we introduce bidirectional interactions between visual and action representations, where scene dynamics guide action generation and motion tokens refine visual reconstruction, enabling mutual reinforcement.

To effectively transfer the learned latent action knowledge into VLA models, we present an effective strategy for knowledge distillation. This strategy enables VLMs to inherit latent action knowledge while preserving its reasoning and instruction-understanding capabilities. It also allows the model to capture rich physical priors, ultimately enhancing generalization and transferability in real-world robotic manipulation. Specifically, we design two types of loss functions: \textit{Latent Action Alignment Loss} and \textit{Reasoning Preservation Loss}. The former transfers physical priors of latent actions from the teacher model (LAM) to the student (VLM) by aligning their latent action representations through both MSE and KL divergence. This helps the student learn physics-aware latent actions that can capture task-relevant motion patterns and future visual changes, allowing the model to rapidly adapt to new manipulation tasks with few shot samples. The latter adopts a next-token prediction objective, enabling the student to generate subtask descriptions based on the current frame and task instruction. It preserves the reasoning and instruction-following abilities of VLMs, ensuring that the distilled model remains robust and generalizable for complex robotic manipulation.


We pre-train the latent action model and perform knowledge distillation on diverse object manipulation datasets encompassing both robot and human hand demonstrations. These datasets include OXE \cite{vuong2023open}, AgiBoT \cite{bu2025agibot}, and EgoDex \cite{hoque2025egodex} that represents the largest and most diverse publicly available collection of dexterous human manipulation data. 
The diversity of scenes and embodiments encourages the model to learn universal latent actions that generalize across visual domains and capture shared task patterns under different scenes, enhancing transferability to downstream robotic tasks. Compared with the latest VLA models \cite{black2024pi_0, intelligence2025pi_}, our approach achieves superior performance in both simulation and real-world environments.

\section{Related Work}
\subsection{Vision-Language-Action Model}

Vision–Language–Action (VLA) models extend Vision–Language Models (VLMs)~\cite{wang2025internvl3, bai2025qwen2} to generate robot actions conditioned on visual observations and language instructions. Early efforts such as RT-1~\cite{brohan2022rt} and Octo~\cite{team2024octo} employ transformer-based policies trained on large-scale collections of robotic trajectories spanning diverse tasks, objects, and environments. RT-2~\cite{zitkovich2023rt} further fine-tunes a pretrained VLM with both vision–language data and robotic demonstrations, discretizing actions into text-like tokens. Following a similar strategy, OpenVLA~\cite{kim2024openvla} adapts the Prismatic VLM~\cite{karamcheti2024prismatic} on the Open X-Embodiment dataset~\cite{o2024open}. Other approaches integrate VLMs with specialized action modules. For instance, RoboFlamingo~\cite{li2023vision} appends a policy head for action prediction, while $\pi_0$~\cite{black2024pi_0} leverages PaliGemma~\cite{beyer2024paligemma} for scene understanding and a flow-matching expert for continuous control. Furthermore, several methods also incorporate goal images~\cite{black2023zero} or video prediction~\cite{hu2024video} as auxiliary tasks to enhance planning and execution. Nevertheless, these methods rely heavily on interactive datasets with ground-truth action labels, which substantially limits the scalability and generalization of VLA models.


\subsection{Latent Action Model}

Recent research \cite{bu2025univla, chen2025villa, ye2024latent} has explored latent actions to address the scalability limitations of VLA models that rely on ground-truth action labels. Latent actions provide compact and transferable representations, enabling learning from large-scale, unlabeled videos. Early studies such as Genie~\cite{bruce2024genie} and LAPO~\cite{schmidt2023learning} introduced unsupervised latent action modeling in video game environments, while DynaMo~\cite{cui2024dynamo} extended this idea with inverse and forward dynamics to learn structured state representations.
In robotic learning, several methods~\cite{ye2024latent, bu2025univla, chen2025villa, zhang2025latent} incorporate latent actions into VLA pretraining, enabling policy learning without explicit action supervision.
For example, LAPA~\cite{ye2024latent} and ViLLA-X~\cite{chen2025villa} extend latent action learning to both human and robot videos, facilitating cross-domain transfer between human demonstrations and robotic executions.
Moto-GPT~\cite{chen2024moto} focuses on motion-centric representation learning by converting videos into discrete motion tokens and co-finetuning them with real robot actions to bridge motion understanding and control.
UniVLA~\cite{bu2025univla} adopts a two-stage pipeline to learn task-centric latent actions, which shows promising results. 
However, existing approaches remain limited by sub-optimal latent representations.
In contrast, we propose to explicitly disentangle latent actions into transferable components (including motion and scene tokens), and  align them with real physical states (e.g., translation and rotation), which makes them easier to distill into downstream robotic tasks.



\section{Approach}
In this section, we introduce the proposed universal latent action learning framework, \textbf{LatBot}, which consists of two key components: Latent Action Disentanglement and Unified Decoder, which are jointly optimized during training. After pre-training the latent action representation, we further distill the learned motion knowledge into VLMs to enhance their action awareness while preserving their original reasoning capability. This enables robot policies to effectively generalize from task reasoning to action execution. 

\begin{figure*}
    \centering
    \includegraphics[width=1.05\linewidth]{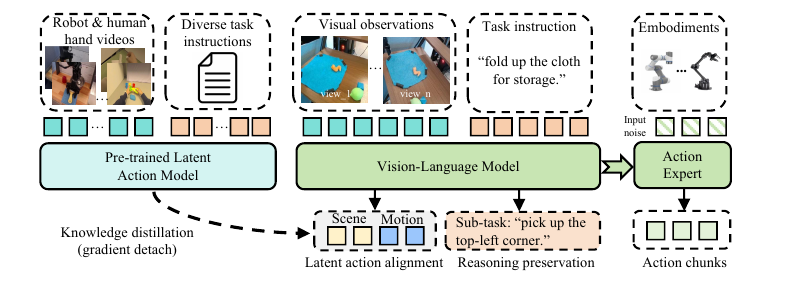}
    \caption{Illustration of the proposed latent action distillation approach for VLA models. By optimizing the VLMs with latent action alignment loss and reasoning preservation loss, we distill generalizable action representations learned from both robot and human hand demonstration videos, while simultaneously maintaining sub-task planning capabilities. This is followed by an action expert module for continuous action prediction.}
    \label{fig:distill}
\end{figure*}


\subsection{Decoupled Latent Action Representation}
Current latent action models predominantly use visual reconstruction as the training objective, which biases them toward learning image-space features rather than motion representations grounded in physical actions \cite{cui2024dynamo, ye2024latent, schmidt2023learning}. As a result, the learned latent actions remain far from executable robot actions, limiting the model’s ability to rapidly adapt to new environments with few samples. Narrowing this gap is essential for establishing a reliable perception–action mapping and achieving efficient transfer to novel scenes. Moreover, existing methods \cite{ye2024latent, chen2025villa} typically entangle all visual variations, including both robot-induced motion and environment-induced changes within a single latent action representation. This entanglement introduces task-irrelevant signals (e.g., background motion or lighting fluctuations), weakens the correspondence between latent actions and true robot dynamics, and ultimately leads to inaccurate action predictions in manipulation tasks.\par

To address these issues, we propose a Universal Latent Action Learning framework that extracts latent actions from multi-frame observations under task-instruction guidance and jointly optimizes them via visual reconstruction and action generation objectives. This design enables the model to acquire physics-related priors (e.g., real-world distances and orientations) that more closely align with executable actions, thereby improving its transferability to downstream robotic manipulation tasks.

Specifically, we propose a \textbf{Decoupled Latent Action} Representation that separates the latent action $Z_{a}$ into two components: the motion representation $Z_{\text{mot}}$, capturing the \emph{active changes} driven by the robot’s own motion, and the scene representation $Z_{\text{sce}}$, capturing the \emph{passive scene changes} induced by environmental dynamics. This decomposition reduces task-irrelevant noise and establishes a clearer correspondence between robot motion, environmental variations, and latent action representations, thereby enhancing performance in downstream manipulation tasks. To extract two types of latent action presentations, we propose to leverage a pretrained vision-language model (VLM) due to its strong contextual understanding to reason about latent actions by integrating visual observations with language instructions. This process can be formulated as follows:
\begin{equation}
    \{Z_{\text{sce}}, Z_{\text{mot}}\} = f_{\text{vlm}}(V_{t:t+k}, \ell),
\end{equation}
where $f_{\text{vlm}}$ denotes the VLM, which takes the visual frames as input from timestep $t$ to $t+k$ along with the task instruction $\ell$. In particular, we introduce two learnable latent action tokens, [\texttt{CP\_SCE}] and [\texttt{CP\_MOT}], into the VLM’s vocabulary, allowing it to encode contextual information into structured scene representations $Z_{\text{sce}}$ and motion representations $Z_{\text{mot}}$, respectively. To fully leverage the VLM’s instruction-following ability for latent action summarization, we design an instruction-tuning template that guides the model in extracting the corresponding latent action representations from multi-frame sequences.

\subsection{Unified Latent Action Decoding}
To ensure that the latent actions focus on the dynamics changes from multiple frames, we further use them as conditional inputs to jointly guide the reconstruction of the future frame and the generation of inter-frame actions. In this way, the visual reconstruction constraint encourages the latent actions to capture observable scene variations, while the action generation objective provides physical-level guidance, enabling the model to establish a closer connection between the latent action and physical motions. Consequently, the model acquires universal latent actions that capture both visual dynamics and physical priors, enhancing prediction across diverse manipulation scenarios and aligning more closely with real robot actions, thereby improving downstream task transferability. 

Specifically, we propose a unified decoder, where scene and motion information are progressively fused through layer-wise bidirectional interactions, which can help the model learn latent actions that are better aligned with real robot dynamics. The decoder is initialized from pretrained image generation model SANA \cite{xie2024sana}, which can leverage the pretrained model’s powerful generative and contextual understanding capabilities.
At each layer of the decoder, the scene and motion representations interact and exchange information, enabling progressive fusion between spatial and dynamic cues. Finally, the future visual frame $V_{t+k}$ and inter-frame actions $A_{t:t+k}$ are decoded based on the fused scene and motion features.
This bidirectional fusion mechanism allows scene dynamics to guide action generation, while motion tokens refine visual reconstruction, enabling mutual reinforcement between the two modalities. 

\begin{table*}[h]
    \centering
    \caption{Comparison of different VLA models across four tasks in two SIMPLER settings on the Google robot.}
    \label{tab:simpler_google}
    \begin{tabular}{c|c|cccc|c}
        \toprule
        \textbf{\shortstack{Google \\ Robot}} & \textbf{Method} & \textbf{\shortstack{Pick \\ Coke Can}} & \textbf{\shortstack{Move \\ Near}} & \textbf{\shortstack{Open/Close \\ Drawer}} & \textbf{\shortstack{Open Top Drawer \\ and Place Apple}} & Avg \\
        
        \midrule
        \multirow{6}{*}{\shortstack{Visual \\ Matching}} 
        & RT-2-X \cite{vuong2023open} & 78.7\% & 77.9\% & 25.0\% & 3.7\% & 46.3\% \\
        & OpenVLA \cite{kim2024openvla} & 18.0\% & 56.3\% & 63.0\% & 0.0\% & 34.3\% \\
        & $\pi_0$ \cite{black2024pi_0} & 87.3\% & 35.0\% & 72.6\% & 16.0\% & 52.7\% \\
        & SpatialVLA \cite{qu2025spatialvla} & 86.0\% & 77.9\% & 57.4\% & 0.0\% & 55.3\% \\
        & RoboVLM \cite{liu2025towards} & 76.3\% & 79.0\% & 44.9\% & 27.8\% & 57.0\% \\
        & villa-X \cite{chen2025villa} & 81.7\% & 55.4\% & 38.4\% & - & - \\
        & DD-VLA \cite{liang2025discrete} & 85.4\% & 67.5\% & 60.6\% & - & - \\
        & MemoryVLA \cite{shi2025memoryvla} & 90.7\% & 88.0\% & 84.7\% & \textbf{47.2\%} & 77.2\% \\
        \rowcolor{gray!20}
        & \textbf{Ours} & \textbf{96.7\%} & \textbf{91.7\%} & \textbf{90.4\%} & 33.3\%  & \textbf{78.0\%}  \\
        \midrule
        \multirow{6}{*}{\shortstack{Variant \\ Aggregation}} 
        & RT-2-X \cite{vuong2023open} & 82.3\% & \textbf{79.2\%} & 35.3\% & 20.6\% & 54.4\% \\
        & OpenVLA \cite{kim2024openvla} & 60.8\% & 67.7\% & 28.3\% & 1.2\% & 39.3\% \\
        & $\pi_0$ \cite{black2024pi_0} & 85.2\% & 40.8\% & 42.1\% & 16.0\% & 46.0\% \\
        & SpatialVLA \cite{qu2025spatialvla} & 88.0\% & 72.7\% & 41.8\% & 6.3\% & 51.8\% \\
        & DD-VLA \cite{liang2025discrete} & 82.5\% & 64.6\% & 23.6\% & - & - \\
        & MemoryVLA \cite{shi2025memoryvla} & 80.5\% & 78.8\% & 53.2\% & \textbf{58.3\%} & 67.7\% \\
        \rowcolor{gray!20}
        & \textbf{Ours} & \textbf{95.7\%}  & 78.3\%  & \textbf{73.0\%}  & 33.3\% & \textbf{70.1\%} \\
        \bottomrule
    \end{tabular}
\end{table*}

\begin{table*}[h]
    \centering
    \caption{Comparison of different VLA models across four tasks in the SIMPLER (Visual Matching) setting on the WidowX robot.}
    \label{tab:simpler_widowx}
    \begin{tabular}{c|cccc|c}
        \toprule
        \textbf{Method} & \textbf{\shortstack{Put Spoon \\ on Towel}} & \textbf{\shortstack{Put Carrot \\ on Plate}} & \textbf{\shortstack{Stack Green Block \\ on Yellow Block}} & \textbf{\shortstack{Put Eggplant \\ in Yellow Basket}} & Avg \\
        
        \midrule
        SpatialVLA \cite{qu2025spatialvla} & 16.7\% & 25.0\% & 29.2\% & \textbf{100\%} & 42.7\% \\
        CogACT \cite{li2024cogact} & 71.7\% & 50.8\% & 15.0\% & 67.5\% & 51.3\% \\
        $\pi_0$ \cite{black2024pi_0} & 62.5\% & 66.7\% & 25.0\% & 12.5\% & 41.7\% \\
        $\pi_{0.5}$ \cite{intelligence2025pi_} & 79.2\% & 58.3\% & 16.7\% & 66.7\% & 55.2\%\\
        villa-X \cite{chen2025villa} & 48.3\% & 24.2\% & 19.2\% & 71.7\% & 40.8\% \\
        UniVLA \cite{bu2025univla} & 52.8\% & 55.6\% & 2.8\% &  80.6\% & 47.9\% \\
        MemoryVLA \cite{shi2025memoryvla} & 75.0\% & 75.0\% & 37.5\% &  \textbf{100\%} & 71.9\% \\
        \rowcolor{gray!20}
        \textbf{Ours} & \textbf{95.8$\%$} & \textbf{87.5$\%$} & \textbf{83.3$\%$} & 83.3$\%$ & \textbf{87.5$\%$} \\
        \bottomrule
    \end{tabular}
\end{table*}


\subsection{Knowledge Distillation for VLA Models}
Although the latent action model (LAM) effectively learns physically grounded latent action representations, its capabilities are limited to scene reconstruction and inter-frame action generation. To bridge this gap and transfer the learned knowledge to the vision–language–action (VLA) model, we propose a latent action knowledge distillation strategy. This approach enables the VLA to inherit motion understanding and physical priors from LAM while preserving its original vision–language reasoning abilities. As a result, the VLA can extract features closer to the action modality and acquire motion planning capabilities, facilitating efficient transfer to downstream manipulation tasks. \par

Specifically, given a pretrained latent action model (denoted as $f_{\text{lam}}$), a language instruction $\ell$, and multiple frames $\{V_t\}_{t=1}^T$, the LAM first extracts latent action representations $Z_a$ conditioned on the instruction:
\begin{equation}
    Z_a = f_{\text{lam}}(\ell, \{V_t\}_{t=1}^T),
\end{equation}
which captures the implicit correspondence between visual dynamics and task semantics. Meanwhile, the vision-language model (VLM) within the VLA model generates its own action representation conditioned only on the first frame $V_1$ and the same task instruction:
\begin{equation}
    \hat{Z}_a = f_{\text{vlm}}(\ell, V_1),
\end{equation}
where $\hat{Z}_a$ is expected to contain future motion information, which is generated by VLMs. 

To align these two types of representations, we design a \textbf{Latent Action Alignment Loss}  $\mathcal{L}_{\text{a}}$ that combines a reconstruction term and a distribution alignment term:
\begin{equation}
    \mathcal{L}_{\text{a}} = \|\hat{Z}_a - Z_a\|_2^2 + \text{KL}\big(p(\hat{Z}_a)\  \|\  p(Z_a)\big),
\end{equation}
where the first term enforces feature consistency and the second encourages distributional alignment, allowing the VLM to gain future frame forecasting capability. Unlike explicit action supervision in VLA models, latent actions are more embodiment-agnostic and naturally align with VLM representations. However, direct alignment may inadvertently compromise the VLM's inherent language understanding and reasoning abilities. To preserve these capabilities, we introduce a \textbf{Reasoning Preservation Loss} $\mathcal{L}_{\text{r}}$ to guide sub-task planning in robot manipulation tasks: 
\begin{equation}
    \mathcal{L}_{\text{r}} = - \sum_{i} \log p(w_{i+1} \mid w_{\le i}, \ell, V_1),
\end{equation}
which preserves the VLM’s reasoning ability and enables it to autoregressively predict the $i+1$ token based on preceding tokens, generating coherent sub-task descriptions conditioned on the current frame and task instruction. Finally, the overall objective for latent action knowledge transfer is defined as:
\begin{equation}
    \mathcal{L} = \mathcal{L}_{\text{a}} + \lambda_{\text{r}} \cdot \mathcal{L}_{\text{r}},
\end{equation}
where $\lambda_{\text{r}}$ balances the trade-off between latent action alignment and reasoning preservation and default to $0.5$ in our experimental settings.

\textbf{Action Expert Finetuning:} after latent action knowledge distillation, the VLM not only retains its original vision–language reasoning capabilities, but also gains the motion planing abilities and generates features that are closely aligned with actions. However, these outputs are still latent representations and not directly executable as robot actions. Therefore, we further perform finetuning in both real-world and simulated robotic environments by incorporating an action expert, enabling precise generation of executable actions. To provide fine-grained supervision for action generation, we decompose the overall action loss into two components:$\mathcal{L}_{ee} + \mathcal{L}_{gripper}$,
where $\mathcal{L}_{ee}$ denotes the loss for the end-effector's translation and rotation, computed using mean squared error. $\mathcal{L}_{gripper}$ denotes the loss for the gripper state, computed using binary cross-entropy to encourage more deterministic behavior.

\section{Experiments}

\begin{table*}[h]
    \centering
    \caption{Comparison of different VLA models on the four LIBERO simulation environments.}
    \label{tab:libero_results}
    \begin{tabular}{c|cccc|c}
        \toprule
        \textbf{Method} & \textbf{LIBERO-Goal} & \textbf{LIBERO-Object} & \textbf{LIBERO-Spatial} & \textbf{LIBERO-Long} & \textbf{Avg} \\
        \midrule
        Diffusion Policy \cite{chi2023diffusion} & 68.3\% & 92.5\% & 78.3\% & 50.5\% & 72.4\% \\
        Octo \cite{team2024octo} & 84.6\% & 85.7\% & 78.9\% & 51.1\% & 75.1\% \\
        OpenVLA \cite{kim2024openvla} & 79.2\% & 88.4\% & 84.7\% & 53.7\% & 76.5\% \\
        TraceVLA \cite{zheng2024tracevla} & 75.1\% & 85.2\% & 84.6\% & 54.1\% & 74.8\% \\
        RDT \cite{liu2024rdt} & 68.2\% & 77.8\% & 60.2\% & 29.0\% & 58.8\% \\
        $\pi_0$ \cite{black2024pi_0} & 94.0\% & 97.8\% & 91.4\% & 85.4\% & 92.2\% \\
        UniVLA \cite{bu2025univla} & 95.6\% & 96.8\% & 96.5\% & 92.0\% & 95.2\% \\
        villa-X \cite{chen2025villa} & 91.5\% & 97.0\% & 97.5\% & 74.5\% & 90.1\% \\
        $\pi_{0.5}$ \cite{intelligence2025pi_} & 98.0\% & 98.2\% & 98.8\% & 92.4\% & 96.9\% \\
        MemoryVLA \cite{shi2025memoryvla} & 96.4\% & 98.4\% & 98.4\% & 93.4\% & 96.5\% \\
        \rowcolor{gray!25}
        \textbf{Ours} & \textbf{98.6\% }& \textbf{98.8\%} & \textbf{99.0\%} & \textbf{95.4\%}  & \textbf{98.0\%} \\

        \bottomrule
    \end{tabular}
\end{table*}
\subsection{Implementation Details}
Our latent action model (LAM) is pre-trained on a combined dataset of OXE \cite{o2024open}, AgiBoT \cite{bu2025agibot}, and the human hand manipulation dataset EgoDex \cite{hoque2025egodex}, encompassing a total of one million video episodes.
For EgoDex, we leverage its rich action annotations, including 3D positions and 6D orientations of both hands, as well as the 3D position of each fingertip, to provide fine-grained supervision for learning high-quality latent actions. Detailed specifications of the action space for both robot and human hand demonstrations are provided in the supplementary materials.
The subsequent knowledge distillation stage is conducted on the same dataset. LAM pretraining and the distillation stage are performed on 16 NVIDIA A100 (40GB) GPUs for 14 and 7 days, respectively. The LAM encoder can be initialized from pretrained vision-language models such as PaliGemma \cite{beyer2024paligemma} or InternVL3.5 \cite{wang2025internvl3}. The unified decoder is initialized from the pretrained image generation model SANA \cite{xie2024sana}. 
We train the model using Fully Sharded Data Parallel (FSDP) with a per-GPU batch size of 16 and a gradient accumulation step of 2, yielding an effective global batch size of 512. By default, LAM operates on 16-frame sequences and represents latent actions using 64 scene representations and 64 action representations. During knowledge distillation and fine-tuning, we use $\pi_{0.5}$ \cite{intelligence2025pi_} as the default VLA backbone.

\subsection{Manipulation Benchmark on SIMPLER}
The SIMPLER \cite{li2024evaluating} benchmark is designed to bridge the real-to-sim gap by recreating realistic scenarios for the Google Robot and WidowX Robot. We evaluate our method on the SIMPLER benchmark and compare it with a wide range of recent VLA models. 
Table \ref{tab:simpler_google} reports results on the Google Robot under the Visual Matching and Variant Aggregation settings. Across both evaluation protocols, our model consistently achieves the best overall performance. Under Visual Matching, our approach reaches 78.0\% average success rate, outperforming all prior open-source models and improving over $\pi_0$ \cite{black2024pi_0} by a significant margin (+25.3\%). Notably, our method also surpasses the closed-source RT-2-X despite using fewer model parameters. Under Variant Aggregation, our model again sets a new state of the art with 70.1\%, exceeding $\pi_0$ by 24.1\% and RT-2-X by 15.7\%. These results demonstrate the robustness of our model across different real-to-sim evaluation settings and its ability to generalize to visually altered scenes. As shown in Table \ref{tab:simpler_widowx}, our method exhibits an even more pronounced advantage on the WidowX robot. It achieves an average success rate of 87.5\%, substantially outperforming all existing VLA models. Compared with the strong baseline $\pi_{0.5}$ \cite{intelligence2025pi_}, our method achieves an improvement of 32.3\%. More importantly, when compared with recent latent-action–based methods such as UniVLA \cite{bu2025univla} (47.9\%) and villa-X \cite{chen2025villa} (40.8\%), our model delivers gains of 39.6\% and 46.7\%, respectively. These results demonstrate that our method effectively transfers latent-action knowledge into the VLA domain, thereby improving the model’s robustness and generalization across diverse tasks.
\begin{figure*}
    \centering
    \includegraphics[width=1\linewidth]{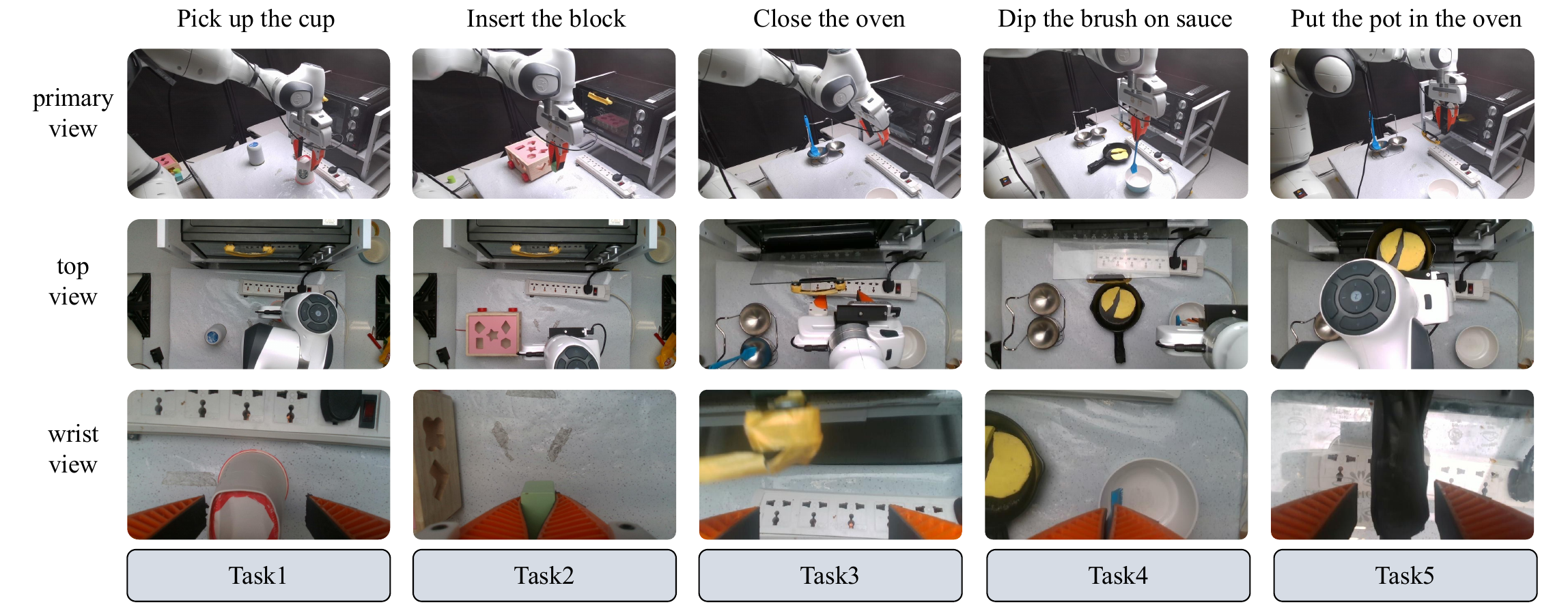}
    \caption{The real-robot Franka setup is equipped with multi-view observations. Tasks include pick-up, insertion, and so on, requiring both translational and rotational motions. In addition, we evaluate more practical scenarios involving interactions with real objects such as thin brushes, heavy frying pans, and real ovens.}
    \label{fig:real_world}
\end{figure*}

\begin{table*}[t]
    \centering
    \caption{Success rate comparison across different tasks under varying the numbers of demonstrations for training.}
    \label{tab:real_world}
    \setlength{\tabcolsep}{4pt}
    \renewcommand{\arraystretch}{1.15}
    \begin{tabular}{l|ccc|ccc|ccc|ccc|ccc|c}
    \toprule
    \multirow{2}{*}{Method} 
    & \multicolumn{3}{c|}{\textbf{Task 1}} 
    & \multicolumn{3}{c|}{\textbf{Task 2}} 
    & \multicolumn{3}{c|}{\textbf{Task 3}} 
    & \multicolumn{3}{c|}{\textbf{Task 4}}
    & \multicolumn{3}{c|}{\textbf{Task 5}} 
    & \multirow{2}{*}{Avg} \\ 
    \cmidrule(lr){2-16}
    & 10 & 50 & full 
    & 10 & 50 & full 
    & 10 & 50 & full 
    & 10 & 50 & full
    & 10 & 50 & full & \\ 
    \midrule
    $\pi_{0}$ \cite{black2024pi_0} & \textcolor{black}{0} & \textcolor{black}{0} & \textcolor{black}{10\%} &
    \textcolor{black}{0} & \textcolor{black}{0} & \textcolor{black}{10\%} & \textcolor{black}{40\%} & \textcolor{black}{40\%} & \textcolor{black}{60\%} & \textcolor{black}{0} & \textcolor{black}{0} & \textcolor{black}{10\%} & \textcolor{black}{0} & \textcolor{black}{0} & \textcolor{black}{20\%} & 12.7\% \\
    $\pi_{0.5}$ \cite{intelligence2025pi_} & 0 & 0 & 20\% & 0 & 0 & 10\% & \textcolor{black}{60\%} & \textcolor{black}{60\%} & \textcolor{black}{80\%} & \textcolor{black}{0} & \textcolor{black}{0} & \textcolor{black}{20\%} & \textcolor{black}{20\%} & \textcolor{black}{20\%} & \textcolor{black}{20\%} & 20.7\% \\
    \rowcolor{gray!25}
    \textbf{Ours} & \textbf{60\%} & \textbf{80\%} & \textbf{60\%} & \textbf{20\%} & \textbf{80\%} & \textbf{40\%} & 
    \textcolor{black}{\textbf{80\%}} & \textcolor{black}{\textbf{100\%}} & \textcolor{black}{\textbf{80\%}} & 
    \textbf{20\%} & \textcolor{black}{\textbf{50\%}} & \textbf{80\%} & \textcolor{black}{\textbf{60\%}} & \textcolor{black}{\textbf{60\%}} & \textcolor{black}{\textbf{80\%}} & \textbf{63.3\%} \\
    \bottomrule
    \end{tabular}
\end{table*}

\subsection{Manipulation Benchmark on LIBERO}

The LIBERO \cite{liu2023libero} benchmark consists of four task suites, which are designed to study lifelong learning in robotic manipulation. We perform experiments on four task suites, each comprising 10 tasks with 50 human-teleoperated demonstrations. Specifically, \textit{LIBERO-Spatial}, \textit{LIBERO-Object} and \textit{LIBERO-Goal} evaluate the understanding of the spatial relationships, object types and different task-oriented behaviors, respectively. \textit{LIBERO-Long} test the ability to generalize the long-horizon tasks with different objects, layouts and goals.
We fine-tune our model on the mixed LIBERO dataset for 60k steps with a batch size of 64. All methods are evaluated over 500 rollouts per task suite (i.e., 50 rollouts per task). As shown in Table \ref{tab:libero_results}, our method achieves the highest overall success rate of 98.0\% across the four LIBERO environments. Compared with the baseline $\pi_{0.5}$, our method achieves a 3.0\% improvement on \textit{LIBERO-Long}, indicating that after latent action knowledge distillation, the VLA acquires stronger motion planning and future-state awareness, which substantially enhances its performance on long-horizon tasks. 

\begin{table*}[h]
    \centering
    \caption{Impact of each component, evaluated in the SIMPLER benchmark. \textbf{UAD}, \textbf{DLA} denotes the unified action decoder, the decoupled latent action representations, respectively.}
    \label{tab:ablation}
    \begin{tabular}{c|cc|cccc|c}
          \toprule
          & UAD  & DLA  & \shortstack{Put Spoon \\ on Towel}  & \shortstack{Put Carrot \\ on Plate} & \shortstack{Stack Green Block \\ on Yellow Block} & \shortstack{Put Eggplant \\ in Yellow Basket} & Avg \\
          \midrule
          UniVLA-style & &  & 41.6\% & 54.2\% & 45.8\% & 62.5\% & 51.0\% \\ 
          Ours-v1 & & \checkmark  & 70.8\% & 66.7\% & 37.5\% & 62.5\% & 59.4\% \\ 
          Ours-v2 & \checkmark &  & 70.8\% & 66.7\% & 41.7\% & 66.7\% & 61.5\% \\ 
          Ours-v3  & \checkmark & \checkmark  & 95.8\% & 87.5\% & 83.3\% & 83.3\% & 87.5\% \\
    \bottomrule
    \end{tabular}
\end{table*}

\subsection{Real-World Evaluation with Franka Robot}

To rigorously evaluate the model’s performance in real-world robot setups, which involve higher uncertainty and demand greater generalization from VLA models, we conduct a series of manipulation experiments using a Franka Research 3 robot with 7 degrees of freedom (DoFs) and a 1-DoF parallel gripper.

\textbf{Task specifications:} 1) To assess the instruction-following and visual understanding abilities, we first design a color discrimination task where the robot must correctly pick up a target cup when both a red and a blue cup are present in its view: \textit{Pick up the cup} (\textbf{Task 1}).
2) To further examine the fine-grained manipulation and physical reasoning capabilities, we introduce four more challenging tasks: \textit{Put the building block into the corresponding slot} (\textbf{Task 2}),  \textit{Close the oven} (\textbf{Task 3}), \textit{Dip the brush in the sauce} (\textbf{Task 4}), and \textit{Put the pot in the oven} (\textbf{Task 5}). 
\textbf{Task 2} and \textbf{Task4} require multi-stage control and precise spatial understanding, while \textbf{Task 3} and \textbf{Task 5} demand delicate gripper control. For example, closing the oven door requires accurately grasping the handle and applying force along the correct motion direction.

Note that each task comprises $100$ demonstrations, collected via human expert teleoperation. To evaluate the model’s few-shot transfer capability, we train it using subsets of $10$, $50$, and all available demonstrations for each task. We compare our approach against $\pi_0$ and $\pi_{0.5}$. Table~\ref{tab:real_world} summarizes the success rates of the models across five real-world manipulation tasks under varying demonstration sizes. Our method consistently outperforms the baselines across nearly all tasks and training settings. For the same number of the training samples (e.g., 50-shot), all models are trained using the same batch size and number of GPUs, with the same amount of training steps and each task is evaluated over 10 trials.

For the color discrimination task (\textbf{Task 1}), our model achieves a 60\% success rate with only 10 demonstrations and 80\% with 50-shot, whereas both baselines fail entirely in the few-shot settings and reach at most 20\% with the full dataset. Interestingly, the 50-shot performance surpasses that of fine-tuning with all available data. This likely results from the full dataset containing redundant action patterns, which cause the latent actions to encode irrelevant variations. Consequently, the VLA model may produce slightly inaccurate actions, reducing overall success. In contrast, the 50-shot subset is more concise and cleaner, allowing latent actions to focus on the core, task-relevant motion features.


For the multi-stage control task (\textbf{Task 2}), where the robot must first pick up a building block and then place it into the corresponding slot, our model demonstrates strong few-shot learning capability, achieving 80\% success with only 50 demonstrations, while the baselines show no success until trained on the full dataset. This task is particularly challenging as it requires both sequential reasoning and precise spatial alignment—the robot must grasp the block accurately and position it correctly in the slot. The strong few-shot performance indicates that our latent actions capture essential task dynamics and structure, enabling the VLA model to plan and execute multi-step manipulations effectively with limited data. 
Similarly, in the fine-grained manipulation tasks (\textbf{Task 3--Task 5}), which demand precise gripper control and spatial reasoning, our method consistently outperforms the baselines. For example, 
\textbf{Task 4} challenges the model with small-object manipulation and fine-grained end-effector control, as the robot must stably grasp a thin handle and execute a targeted dipping motion without disturbing the bowl. In this task, our method achieves a 50\% success rate with only 50 demonstrations, while $\pi_{0.5}$ fails to complete the task. When using the full dataset, our approach surpasses $\pi_{0.5}$ by 60\%. \textbf{Task 5} requires precisely grasping the center of the pan handle, otherwise the task will fail. Our method outperforms $\pi_{0.5}$ in both the few-shot and full-shot
settings. This demonstrates that latent action learning effectively equips the model with structured motion priors and fine-grained physical understanding, enabling stable small-object manipulation even under few-shot settings.

\subsection{Components Analysis}
The ablation results in Table \ref{tab:ablation} show that both the decoupled latent action representations (DLA) and the unified action decoder (UAD) play essential and complementary roles in improving manipulation performance. Starting from the UniVLA-style baseline of 51.0\%, introducing DLA alone yields a clear gain to 59.4\% by isolating motion-critical cues from irrelevant environment changes, allowing the model to form cleaner and structured latent actions. Adding only UAD  further improves the average to 61.5\%, as the decoder strengthens the mapping between latent actions and executable robot actions, reducing modality gap during action generation. When both components are combined, the model achieves a substantial jump to 87.5\%, consistently outperforming all other variants across every task. This strong synergy arises because DLA provides structured, manipulation-relevant latent actions, while UAD injects physical priors into the latent action learning process, enabling the robot to generate more accurate and physically consistent action predictions.

\section{Conclusion}

In summary, we propose a universal latent action learning framework, \textbf{LatBot}, and demonstrates that learning transferable latent actions from large-scale object manipulation videos (e.g., robot and human hand), substantially enhances generalization in downstream robotic tasks. By integrating task instructions with multi-frame observations, jointly optimizing future frame reconstruction and action sequence prediction, and disentangling latent actions into motion and scene tokens, our framework effectively captures rich physical priors while filtering out irrelevant dynamics. Experiments show that distilling these latent actions into VLA models yields strong performance across both simulated and real-world robotic platforms. Notably, even with only a few real-world demonstrations on a Franka robot, our method shows that latent actions offer a robust, generalizable representation for complex manipulation tasks, including pick-and-place of thin objects (e.g., brushes) and precise block insertion requiring fine-grained motions. 

These results highlight a key insight: explicitly incorporating physical priors and disentangling motion from environmental changes significantly enhances the transferability of learned latent action representations. For future work, we aim to extract additional latent tokens from larger and more diverse manipulation video datasets, further scaling VLA models and exploring their potential for more complex, long-horizon, and multi-embodiment robotic tasks.

\section*{Acknowledgement}
This work was supported in part by Science and Technology Innovation (STI) 2030---Major Projects under Grant 2022ZD0208700, and National Natural Science Foundation of China under Grant 62376264.
{
    \small
    \bibliographystyle{ieeenat_fullname}
    \bibliography{main}
}

\clearpage
\setcounter{page}{1}
\maketitlesupplementary

\section{Implementation Details}
\label{sec:details}
\subsection{Dataset Details}
We pre-train the latent action model using a combination of robot and human manipulation data, with a total of 1 million video episodes.
For robotic manipulation data, we use the OXE \cite{o2024open}, DROID \cite{khazatsky2024droid}, and AgiBoT \cite{bu2025agibot} datasets. For human hand manipulation, we utilize EgoDex \cite{hoque2025egodex}, which provides detailed hand pose annotations and represents the largest and most diverse dataset for dexterous human manipulation to date. EgoDex provides full bimanual skeletal joints, where each action at a time step is represented by the 3D position of each wrist, the 6D wrist orientation, and the 3D positions of the five fingertips on each hand, resulting in a 48-dimensional action vector. To unify the action representations of robots and human hands during latent action pretraining, we design a \textbf{Unified Action Space} with a total dimensionality of 44:
\begin{itemize}
    \item \textbf{Dimensions 1--7 (Left hand / left arm):}
    Changes in $x$, $y$, $z$, Euler orientation, and gripper state.
    \item \textbf{Dimensions 8--14 (Right hand / right arm):}
    The same set of changes for the right hand or right robotic arm.
    \item \textbf{Dimensions 15--44 (Bimanual fingertips):}
    3D position changes of the ten fingertips (five for each hand), totaling 30 dimensions.
\end{itemize}
We additionally define a \textbf{Unified State Space} for both robots and human hands:
\begin{itemize}
    \item \textbf{Dimensions 1--8 (Left hand / left arm):}
    Current-time $x$, $y$, $z$, quaternion orientation (4D), and gripper state.
    \item \textbf{Dimensions 9--16 (Right hand / right arm):}
    The same information for the right hand or right robotic arm.
    \item \textbf{Dimensions 17--46 (Bimanual fingertips states):}
    3D positions of ten fingertips from both hands at the current time step.
\end{itemize}
Since EgoDex provides 6D orientations, we convert them to Euler angles for the unified action space and to quaternions for the unified state space. The detailed composition of the datasets and mixture weights are listed in Table. \ref{tab:dataset_mix}.

\begin{table}[t]
    \caption{Mixture of datasets used during pretraining, including OXE \cite{o2024open}, DROID \cite{khazatsky2024droid}, and EgoDex \cite{hoque2025egodex}.}
    \label{tab:dataset_mix}
    \centering
    \begin{tabular}{l r}
        \toprule
        \textbf{Dataset Name} & \textbf{Ratio} \\
        \midrule
        Fractal \cite{brohan2022rt} & 12.8\% \\
        Kuka \cite{kalashnikov2018scalable} & 12.8\% \\
        Bridge \cite{walke2023bridgedata} & 11.8\% \\
        Taco Play \cite{mees2022grounding} & 2.7\% \\
        Jaco Play \cite{dass2023jacoplay} & 0.4\% \\
        Berkeley Cable Routing \cite{luo2024multistage} & 0.2\% \\
        Roboturk \cite{mandlekar2019scaling} & 2.1\% \\
        Viola \cite{zhu2023viola} & 0.8\% \\
        Berkeley Autolab UR5 \cite{BerkeleyUR5Website} & 1.1\% \\
        Toto \cite{zhou2023train} & 1.8\% \\
        Language Table \cite{lynch2023interactive} & 4.4\% \\
        Stanford Hydra Dataset \cite{belkhale2023hydra} & 4.6\% \\
        Austin Buds Dataset \cite{zhu2022bottom} & 0.2\% \\
        NYU Franka Play Dataset \cite{cui2022play} & 0.6\% \\
        Furniture Bench Dataset \cite{heo2025furniturebench} & 2.5\% \\
        UCSD Kitchen Dataset \cite{ucsd_kitchens} & $<0.1\%$ \\
        Austin Sailor Dataset \cite{nasiriany2022learning} & 2.2\% \\
        Austin Sirius Dataset \cite{liu2025robot} & 1.7\% \\
        DLR EDAN Shared Control \cite{quere2020shared} & $<0.1\%$ \\
        IAMLab CMU Pickup Insert \cite{saxena2023multi} & 0.9\% \\
        UTAustin Mutex \cite{shah2023mutex} & 2.2\% \\
        Berkeley Fanuc Manipulation \cite{zhu2023fanuc} & 7.8\% \\
        CMU Stretch \cite{nasiriany2022learning} & 1.5\% \\
        BC-Z \cite{jang2022bc} & 6.8\% \\
        FMB Dataset \cite{luo2025fmb} & 7.1\% \\
        DobbE \cite{shafiullah2023bringing} & 1.4\% \\
        DROID \cite{khazatsky2024droid} & $<0.1\%$ \\
        AgiBoT-$\alpha$ \cite{bu2025agibot} & 6.3\% \\
        EgoDex \cite{hoque2025egodex} & 11.1\% \\
        \bottomrule
    \end{tabular}
\end{table}

\begin{figure*}[h]
    \centering
    \includegraphics[width=1\linewidth]{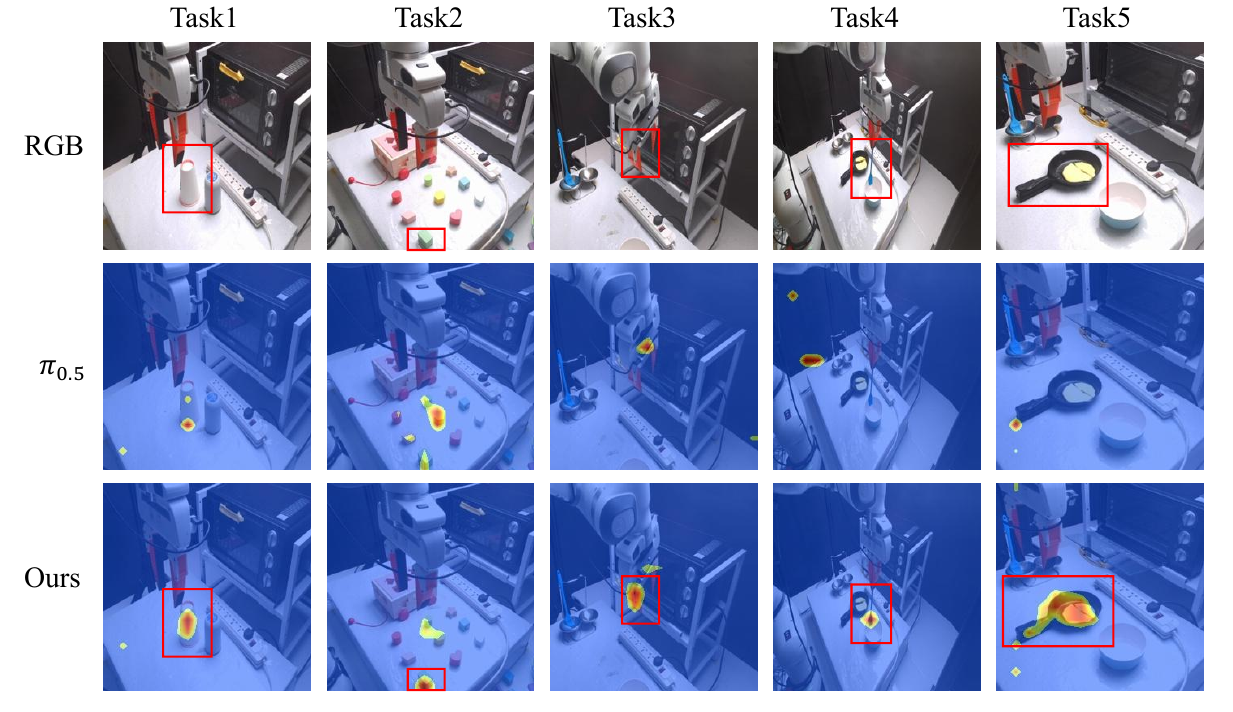}
    \caption{Analysis of the effects the latent-action distillation on the VLM. Following \cite{kang2025your}, we visualize the attention maps between the final text token and the visual features for both $\pi_{0.5}$ and our model across different real-robot tasks. The red bounding boxes in the RGB images mark the task-specific target, while the red boxes on our model’s attention maps highlight the regions with the strongest activations.
    The results show that after latent-action distillation, the VLM of our model exhibits enhanced spatial grounding capabilities, with its attention maps consistently concentrated within the red box.}
    \label{fig:effect_latent_action}
\end{figure*}

\subsection{Model Details}
Our latent action model employs a vision-language model (VLM) as the latent action encoder, which summarizes the inter-frame dynamics into a sequence of latent action representations under language guidance. This VLM component can be any pretrained vision-language model, such as PaliGemma \cite{beyer2024paligemma} or InternVL \cite{wang2025internvl3}. In this work, we default to using InternVL3.5-2B. For the latent action decoder, we adopt an architecture similar to SANA-1.6B  \cite{xie2024sana} and initialize it with the corresponding pretrained weights. The instruction-tuning template used to summarize the latent action representations, including both the question and answer components, is defined as follows:
\begin{tcolorbox}[
    colback=gray!10,      
    colframe=gray!60,    
    arc=3mm,             
    boxrule=0.5pt,       
    left=3mm, right=3mm, top=2mm, bottom=2mm, 
]
    \textit{Given the instruction ``\{sent\}'' and the video frames, reason about what happens within this time span. Explain how the overall scene changes, and identify the temporal dependencies between consecutive frames. Highlight the actions, interactions, and transitions that drive the scene’s evolution. Answer: Scene evolution description: [\texttt{CP\_SCE}]. Action dynamics description: [\texttt{CP\_MOT}].}
\end{tcolorbox}
During the knowledge distillation stage, we use the VLM from $\pi_{0.5}$ as our student model and the teacher model is the pretrained latent action model.
\subsection{Training Details}
During the latent action pretraining stage, we first train the model on a mixed dataset containing both robot and human manipulation data. The pretraining runs for 14 days on 16 RTX A100 GPUs (40GB), with a per-GPU batch size of 16 and a gradient accumulation step of 2, resulting in a total batch size of 512. In this stage, all parameters of the latent action model are optimized except for the vision encoder of the VLM, which remains frozen. We use the AdamW optimizer with $\beta_1 = 0.9$, $\beta_2 = 0.95$. The learning rate is initialized at $1.0 \times 10^{-4}$, followed by a 2,000-step warm-up phase and a cosine decay schedule that gradually anneals it to a minimum of $2.5 \times 10^{-6}$.

For the latent action distillation stage, we use the same dataset as in pretraining. We jointly fine-tune all parameters of the student model (VLM from the vision-language-action model) while keeping the teacher model (VLM from the latent action model) frozen. The learning rate schedule follows the same configuration as in pretraining. The distillation process lasts 7 days on 16 RTX A100 GPUs (40GB), with a per-GPU batch size of 8 and a gradient accumulation step of 2, yielding a total batch size of 256. During both simulation and real-robot fine-tuning, we jointly fine-tune the vision encoder, the VLM, and the action expert components of the VLA. Following \cite{intelligence2025pi_}, we apply quantile normalization for action and state normalization.
\section{Additional Analysis}

\textbf{The effect of the latent action on the VLM.} In real-robot experiments, we observe that our model exhibits strong spatial understanding, enabling it to accurately place blocks into their corresponding slots. To further examine the effect of latent action knowledge on the VLM, we analyze the visual grounding capability of the model before and after distillation. Since LLMs decode in an auto-regressive manner, information gradually accumulates from earlier tokens to later ones, causing the final text token to incorporate the semantic context of the entire instruction \cite{kang2025your}. Therefore, the query vector of the last input text token serves as a representative probe for evaluating sentence-level grounding. We use the query vector of the last input text token as a sentence-level representation for computing attention over image features. 

Specifically, given the query token, we extract the attention weights from the query to all image tokens across all layers and heads. For each attention head, we take the first $P^2$ entries and reshape them into a spatial attention map with size of $P \times P$, where $P$ denotes the patch size. The attention map is then binarized by assigning value 1 to elements above the mean and 0 otherwise. Next, we detect connected components $\{C_i\}_{i=1}^N$ using 8-neighborhood connectivity and compute the spatial entropy:
\begin{equation}
    H = - \sum_{i=1}^{N} P(C_i) \log P(C_i),    
\end{equation}
where $P(C_i) = \frac{|C_i|}{\sum_{j=1}^{N} |C_j|}$. An attention map is considered more spatially localized when it exhibits lower spatial entropy. For visualization, we report the attention map with the lowest spatial entropy among all layers and heads, as it best captures the model's most concentrated grounding behavior.

As shown in Fig. \ref{fig:effect_latent_action}, we compare the attention maps of the last text token over image features between $\pi_{0.5}$ and our method across various real-world robotic tasks. Results show that after latent-action distillation, the VLM can localize task-relevant targets more accurately based on the instruction. When distractors are present (Task2), it exhibits an even stronger response to the true target (reflected by darker attention regions).

\end{document}